\documentclass[journal]{misc/vgtc}                     %

\usepackage{wrapfig}
\usepackage{multirow}
\usepackage{enumitem}
\usepackage{amsmath}%
\usepackage[pagebackref,bookmarks]{hyperref}
\usepackage[nameinlink,capitalise]{cleveref}
\usepackage{booktabs}                  %
\usepackage{mathptmx}   
\usepackage{xspace}
\usepackage{amsfonts}
\usepackage{caption}

\usepackage[svgnames]{xcolor}
\usepackage{todonotes}
\usepackage{setspace}
\usepackage{soul}

\definecolor{slateblue}{rgb}{0.42, 0.35, 0.8}
\definecolor{skyblue}{rgb}{0.53, 0.81, 0.92}
\definecolor{steelblue}{rgb}{0.27, 0.51, 0.71}
\newcommand{\re}[2]{#1}

\newcommand{\system}[1]{\texttt{LLM Analyzer}} %

\newcommand{\furuiglyph}[1]{\raisebox{-1mm}{\includegraphics[height=3.5mm]{figs/glyph-#1.pdf}}\xspace}

\newcounter{textexamplecounter}
\crefname{textexamplecounter}{\protect{Example}}{\protect{examples}}
\Crefname{textexamplecounter}{\protect{Example}}{\protect{Examples}}

\newcommand{\textexamplenofig}[1]{
    \vspace{2mm}
    
    \noindent
    \begin{minipage}{\linewidth}
    \colorbox{gray!10}{
        \begin{minipage}{0.965\linewidth}
        \raggedright
        \vspace{2mm}
        \setlength{\parindent}{0cm}
        {
            \fontsize{0.8em}{0.5em}\selectfont
            \it #1
            
        }
        \vspace{1mm}
        \end{minipage}
    }
    \end{minipage}
}

\onlineid{0}

\vgtccategory{Research}

\vgtcpapertype{please specify}

\title{Understanding Large Language Model Behaviors through \\ Interactive Counterfactual Generation and Analysis}

\author{%
  Furui Cheng,
  Vilém Zouhar,
  Robin Shing Moon Chan,
  Daniel Fürst,
  Hendrik Strobelt,
  Mennatallah El-Assady
}

\authorfooter{
  \item Furui Cheng, Vilém Zouhar, Robin Shing Moon Chan, and Mennatallah El-Assady are with ETH Zürich. Email: \{furui.cheng, vilem.zouhar, robin.chan\}@inf.ethz.ch, menna.elassady@ai.ethz.ch,
  \item Daniel Fürst is with the University of Konstanz. \\Email: daniel.fuerst@uni-konstanz.de,
  \item Hendrik Strobelt is with IBM Research. E-mail: hendrik.strobelt@ibm.com.
}

\abstract{%
Understanding the behavior of large language models (LLMs) is crucial for ensuring their safe and reliable use. However, existing explainable AI (XAI) methods for LLMs primarily rely on word-level explanations, which are often computationally inefficient and misaligned with human reasoning processes. Moreover, these methods often treat explanation as a one-time output, overlooking its inherently interactive and iterative nature. In this paper, we present \system{}, an interactive visualization system that addresses these limitations by enabling intuitive and efficient exploration of LLM behaviors through counterfactual analysis. Our system features a novel algorithm that generates fluent and semantically meaningful counterfactuals via targeted removal and replacement operations at user-defined levels of granularity. These counterfactuals are used to compute feature attribution scores, which are then integrated with concrete examples in a table-based visualization, supporting dynamic analysis of model behavior. A user study with LLM practitioners and interviews with experts demonstrate the system’s usability and effectiveness, emphasizing the importance of involving humans in the explanation process as active participants rather than passive recipients.
}


\teaser{
\centering
\includegraphics[width=\linewidth, alt={A view of a city with buildings peeking out of the clouds.}]{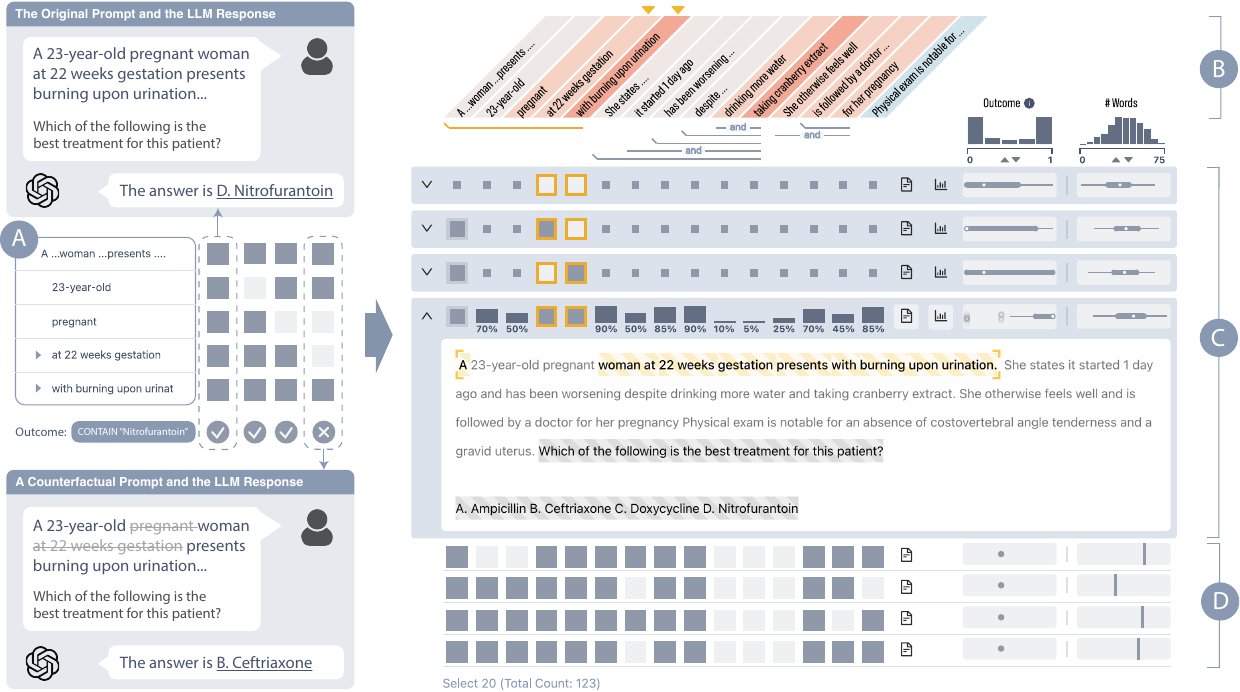}
\caption{%
  \system{} enables users to analyze and understand LLM behaviors through meaningful  counterfactuals.
  From the user's input prototype sentences, such as a medical question in this example, the system generates meaningful segments for perturbation.
  (A) Users can interactively adjust their granularities and specify alternative segments for replacements, which are then used by the system to create meaningful counterfactuals. 
  The system enables users to analyze the LLM  by inspecting and interactively aggregating the counterfactual examples in a table-based visualization.
  (B) The table header shows the segments' text, dependencies, and feature attributions.
  (C) By grouping the counterfactuals by segments of interest, users can assess their joint influence on the predictions.
  (D) In addition to the statistical results, users could view concrete examples to understand the LLM's prediction in alternative scenarios. 
}
\label{fig:teaser}
}

\graphicspath{{figs/}{figures/}{pictures/}{images/}{misc/}{./}} %

\begin{document}

\maketitle

\section{Introduction}

Large Language Models (LLMs) have shown remarkable capabilities in interpreting textual instructions~\cite{ouyang2022training} and solving complex tasks~\cite{thirunavukarasu2023trialling}.
As their adoption grows across a wide range of applications, understanding how and why these models generate specific outputs becomes increasingly critical for ensuring their safety and reliability.
To this end, developing methods that enhance transparency and help users better understand the behavior and limitations of the model is essential.

\smallskip

Existing eXplainable Artificial Intelligence (XAI) approaches for interpreting local model behavior, such as feature attribution methods~\cite{ribeiro2018lime}, primarily provide static, one-shot explanations. 
While these methods can be informative in certain contexts, they suffer from two major limitations. 
First, explanation is inherently an interactive and iterative process: users often seek to ask follow-up questions, test new hypotheses about model behavior, and progressively refine their understanding~\cite{miller2019explanation}. 
Static and monolithic explanations do not accommodate this natural exploratory workflow. 
Second, these methods typically operate at the word level, such as quantifying the influence of individual tokens on the model’s prediction. 
However, word-level representations cause unnecessarily long running time for certain algorithms, including most of the commonly-used removal-based methods~\cite{lundberg2017shap, ribeiro2018lime} and fail to capture the semantic units that humans use for reasoning. 
Humans generally interpret and explain decisions in terms of higher-level meanings, such as phrases, propositions, or claims, rather than isolated words.

\smallskip

To address these limitations, we propose \system{}, an interactive visualization system for understanding LLM behavior. 
Rather than automatically generating and presenting static explanations, \system{} empowers users to actively engage in the analysis process by exploring the \textit{counterfactuals}-variations of a target input crafted to understand the model's prediction under a what-if scenario. 
By constructing and analyzing these counterfactuals, users can directly observe how the model’s outputs change in response to specific input modifications, test their own hypotheses, and iteratively refine their mental model of the LLM’s behavior through hands-on exploration.

\smallskip

The system is powered by a novel text segmentation technique that leverages the dependency structure of the text to segment it into semantically coherent units and organize them hierarchically.
This interpretable representation enables the generation of meaningful counterfactuals by selectively removing or replacing specific segments. 
By evaluating model predictions on these counterfactuals, the system quantifies the influence of each input component on the model’s prediction through the KernelSHAP aggregation~\cite{lundberg2017shap}. 
The resulting attribution scores are visualized to support user exploration and understanding.

\smallskip

We evaluate the usability and usefulness of \system{} through a user study. 
In the user study, the participants are asked to answer explanatory questions, \textit{Why (not)}, \textit{How to be that}, and \textit{How to still be}, by analyzing an LLM's prediction in a multi-hop explanation question, using our system. 
Overall, they all completed the tasks and provided positive feedback regarding both usability and usefulness, indicating that the system effectively supports the proposed workflow and aids users in understanding LLM behavior. 
Additionally, we conduct expert interview with experienced XAI and natural language processing (NLP) researchers. 
The qualitative feedback from both expert and non-expert users offers insights into the system's usage and limitations.
To sum up, the major contributions of this paper include:

\begin{itemize}[topsep=1mm]
\item \system{}, an interactive visualization tool to support LLM practitioners and domain experts in understanding LLM behaviors by analyzing meaningful counterfactuals.
\item A time-efficient algorithm for generating grammatically correct and syntactic-structure-preserving counterfactuals via removing and replacing text segments in user-defined granularity.
\item A user study and expert interviews, providing useful insights into the usage of the system in understanding LLM behaviors.  

\end{itemize}

\section{Research Questions and Challenges}
\label{sec:questions}

We explore the question and general user needs related to understanding LLM behavior without anchoring them to specific application scenarios. 
By reviewing current approaches and their limitations, we summarize two key challenges in enabling efficient LLM behavior analysis. 

\subsection{User Needs in Model Behavior Understanding}

\re{In this paper, we address the challenge of explainability in a general context. We view explainability as a means to help users develop accurate mental models of how a machine learning model behaves~\cite{liao2021human}, enabling them to evaluate whether its behavior aligns with their knowledge and values~\cite{doshi2017towards}. Misalignments can reveal potential model biases and flaws. Our target users are LLM developers and domain experts who seek to understand and assess model behaviors.
}{In this paper, we address the problem of explainability in a general context. We consider the needs of both LLM developers—who aim to assess, debug, and improve model behavior—and LLM users—who seek to justify model outputs in support of their decisions.}

To ground these user needs, we adopt a question-driven design approach. 
Specifically, we draw on the XAI Question Bank, a framework developed by Liao et al.~\cite{liao2020questioning}, which compiles potential user questions about model behavior based on empirical studies. 
The XAI Question Bank organizes these questions into ten categories; four focus on understanding a model’s local behavior, which fall within the scope of this paper. 
We present these four categories along with rephrased versions of their representative questions, adapted to the context of LLMs. In practical scenarios, users may only be interested in a subset of these questions, depending on their specific goals and tasks.
Nevertheless, we consider all four categories to provide a comprehensive foundation for designing explanation interfaces that support broad and flexible assessment of LLM behavior across diverse use cases.

\begin{itemize}[leftmargin=*]
    \setlength\itemsep{0.1em}
    \item \textbf{Why} and \textbf{Why not}: 
    \textit{Why does the model (not) make this prediction?}
    \item \textbf{What-if}: 
    \textit{What would the model predict if the text changes to ...? }
    \item \textbf{How to be that}: 
    \textit{How to get a different prediction by making a (minimal) change to the text?}
    \item \textbf{How to still be this}: 
    \textit{What is the scope of change permitted to still get the same prediction? }
\end{itemize}

The \textit{Why} and \textit{Why not} questions seek to identify the features that influence a model’s prediction. 
In the context of language models, these features refer to specific parts of the input text or their underlying semantics that contribute to the output. 
The \textit{What-if} question allows users to test hypotheses by examining how specific input changes affect the model's response. 
The \textit{How to be that} question explores counterfactual scenarios in which the model would produce a different prediction, while the \textit{How to still be this} question requires more nuanced explanations by identifying which elements are essential for preserving the current prediction. 
We use these four question types to guide the design of our system, with the goal of enabling users to efficiently explore and answer them in the context of LLM behavior to gain a comprehensive understanding of LLM behaviors.

\subsection{Current Local Explanation Approaches}

While answering \textbf{What-if} questions typically involves directly modifying the input and observing the resulting prediction~\cite{wexler2020whatif}, the other question types can be addressed using explainable AI (XAI) methods. In this section, we review the most commonly used types of XAI techniques relevant to understanding model behavior.

\smallskip

\begin{wrapfigure}{l}{0.2\linewidth}
\begin{center}
    \vspace{-2em}
    \includegraphics[width=\linewidth]{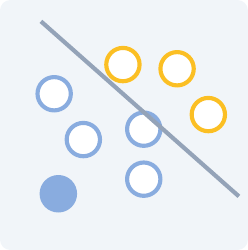}
    \vspace{-3em}
\end{center}
\end{wrapfigure}
\textit{Feature attribution} methods use a (linear) additive model to describe an ML model's local behaviors and quantify the influences of each feature to the prediction~\cite{lundberg2017shap}, which helps answer the \textit{Why} and \textit{Why not} questions. 
Among all attribution methods, removal-based methods~\cite{covert2021explaining}, represented by LIME~\cite{ribeiro2018lime} and KernelSHAP~\cite{lundberg2017shap}, are the most widely used because they are \textit{model-agnostic}-requiring no assumption of model structure and applicable to any ML model. 

\smallskip

\begin{wrapfigure}{l}{0.2\linewidth}
\begin{center}
    \vspace{-2em}
    \includegraphics[width=\linewidth]{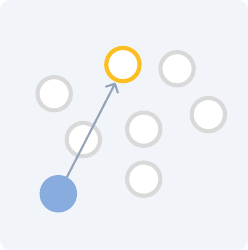}
    \vspace{-3em}
\end{center}
\end{wrapfigure}
\textit{Counterfactual explanations}~\cite{wachter2017counterfactual} are one of the most commonly used example-based explanation methods. They are defined as counterfactuals with a minimal difference from the original instance, leading to a different prediction.    
In addition to the \textit{Why} and \textit{Why not} questions, counterfactual explanations help users answer the \textit{How to be that} question by suggesting the minimal changes required to alter the prediction, e.g., ``\textit{After increasing your monthly income by \$500, the credit assessment model will change the prediction to positive and accept your request.}''

\smallskip

\begin{wrapfigure}{l}{0.2\linewidth}
\begin{center}
    \vspace{-2em}
    \includegraphics[width=\linewidth]{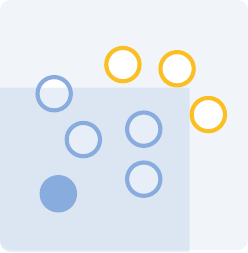}
    \vspace{-3em}
\end{center}
\end{wrapfigure}
The \textit{Anchor}~\cite{ribeiro2018anchors} method uses rules to find the sufficient conditions for model prediction, e.g., if certain features' values are fixed, no matter how other features change, the model will always give the same prediction. 
In addition to the \textit{Why} and \textit{Why not} questions, anchors answer the \textit{How to still be this} question. 
Compared with additive feature attributions, Anchor explanations also communicate features' joint influence on the prediction that could not be modeled using a linear model.

\subsection{Challenge-I: Explanation is an Interactive Process}
\label{sec:challenge-I}

The aforementioned methods provide partial insights into the four core types of explanatory questions, and they position users as passive recipients of the unified, static explanations.
However, explanation is inherently an interactive and iterative process~\cite{miller2019explanation}, in which users actively seek to make sense of a system’s behavior based on their own goals, expectations, and prior knowledge.
Users often formulate hypotheses about how the model works and how the model should work.
An effective explanation framework should support the iterative refinement and validation of these hypotheses. 

\smallskip

Here are some illustration examples. 
In one case, a model developer seeks to evaluate whether a hiring recommendation model exhibits bias. Upon discovering that input segments revealing an applicant’s gender receive high attribution scores in the model’s prediction (\textit{Why}), the developer may hypothesize that the model is incorporating gender information and potentially discriminating against certain groups. To gather more concrete evidence, they further investigate how altering gender-related content affects the model’s output (\textit{What-if}) and under what conditions such information influences the decision.
In another example, a user aims to determine whether the model's behavior aligns with their domain knowledge. They examine whether specific input segments alone are sufficient to produce a given prediction (\textit{How to still be this}). If those segments are found insufficient, the user searches for counterfactual explanations, instances that yield different predictions despite minimal changes to the input (\textit{How to be that}).

\smallskip

Providing users with an efficient workflow with well-integrated explanations remains a general challenge in XAI.
In this paper, we investigate its solution in the context of understanding LLM behaviors. 

\subsection{Challenge-II: Define Interpretable Data Representation}
A preliminary step in generating explanations is constructing a simplified representation of the input~\cite{ribeiro2018lime, lundberg2017shap}. These simplified inputs are typically binary vectors, denoted as $z \in \{0,1\}^{d}$, where each dimension indicates the presence or absence of a specific component in the input. Ribeiro et al.~\cite{ribeiro2018lime} emphasized that these components should be human-interpretable. For instance, in image data, interpretable components might correspond to contiguous pixel regions (i.e., superpixels), where the binary vector encodes whether each superpixel is included or excluded.
This representation allows for systematic perturbation of the input by selectively removing or modifying components to assess their influence on the model's prediction. Constructing such a simplified and interpretable representation involves identifying meaningful components within the input and defining mapping functions between the original input and its simplified form.

\smallskip

In the case of text data, existing approaches often treat individual words as interpretable components, representing the input as a binary vector indicating the presence or absence of specific words~\cite{ribeiro2018lime, lundberg2017shap}. However, word-level interpretable representations come with several limitations.
First, in long-form text, the number of features (i.e., words) becomes large, requiring significantly more samples for analysis and leading to increased computational cost.
Second, arbitrary perturbations at the word level often produce unnatural or ungrammatical sentences, resulting in out-of-distribution inputs that can compromise the validity and interpretability of the explanation.
Third, word-level explanations often do not align with how humans reason—people tend to interpret text based on higher-level semantic units, such as the statements in the text. As a result, word-level explanations can be unintuitive and less useful for supporting human understanding.
Defining interpretable representations of text with a reduced number of components while aligning with human reasoning remains a fundamental challenge in explaining the behavior of language models.

\section{Design LLM Explanations as an interactive process}

We build LLM Explanations as an interactive process that allows users to propose new hypotheses and validate them through tailored and context-specific explanations to answer the four questions.
To ground our approach, we first examine how the three major types of model-agnostic explanations are commonly computed. 
As demonstrated in the removal-based explanation framework proposed by Covert et al.~\cite{covert2021explaining}, these methods typically operate by perturbing the original input, often by simulating feature removal, to generate counterfactuals. The model’s predictions on these perturbed instances are then aggregated to identify which features most influence its decisions.

\smallskip

\begin{figure}[htb]
    \centering
    \includegraphics[width=\linewidth]{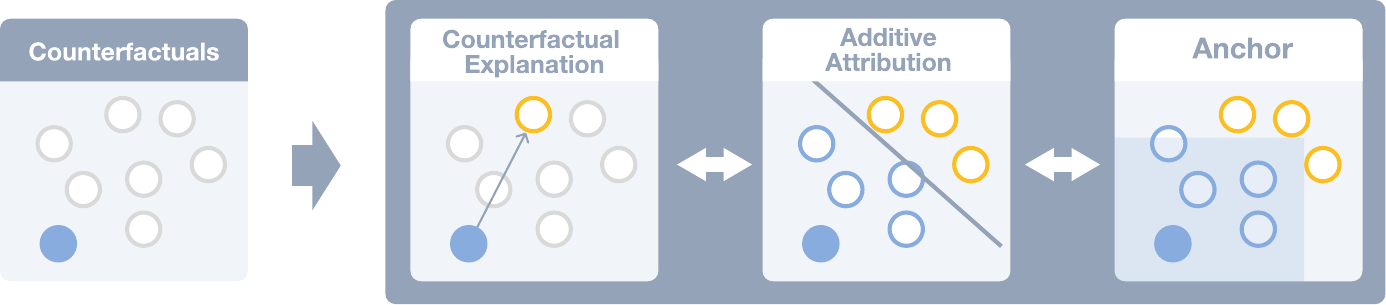}
    \caption{
    Users interactively explore the LLM's responses to the counterfactuals to derive different types of explanations, thereby fostering a more comprehensive understanding of the model’s behavior.
    }
    \label{fig:problem-2}
\end{figure}

This insight motivated a shift in our design: rather than automatically generating and displaying fixed explanations, we enable users to take an active role in the \textit{counterfactual generation} and \textit{analysis} processes~(\Cref{fig:problem-2}). 
With the ability to construct and explore the counterfactuals, 
users can directly observe how the model responds to changes in the input, 
test their hypotheses in answering the four questions, 
and iteratively refine their mental model of the LLM’s behavior.

\subsection{Workflow and User Tasks}
\label{sec:requirements}

We concretize the design problems by discussing the analytical tasks in counterfactual-assisted LLM analysis.
In this study, we aim to support both LLM users and practitioners in analyzing LLMs. 
Based on the discussions in prior sections, we propose an analytical workflow composed of the following user tasks.

\begin{enumerate}[label=\textit{T\arabic*}]
    \item \label{t:create-exp}
    \textit{Create and customize batches of meaningful counterfactuals.} 
    The starting point of conducting a counterfactual-based analysis is to define the scope of the desired counterfactuals, which includes selecting a proper granularity (e.g., words, phrases, sentences) to perform perturbations and customizing the perturbation rules, e.g., fixing some sentences and \re{replacing}{replacement} some segments with alternatives. 
    \re{Afterwards}{Afterward}, users apply the tool to sample batches of meaningful counterfactuals for further analysis.

    \item \label{t:explore-cfs}
    \textit{Explore the counterfactual collection.} 
    After getting the counterfactual collections and corresponding LLM responses, users need to gain an overall understanding of this data collection by interactive exploration. 
    Users may want to locate and inspect concrete counterfactual examples during the explorations to understand how specific perturbations may influence the LLM response. 

    \item \label{t:attribution}
    \textit{Understand how individual segments influence the generation.} 
    After the exploration, users investigate the feature attribution (or the feature-outcome correlations) calculated using the counterfactuals to identify important text segments. Users compare the additive explanations with their assumptions from prior knowledge, where the misalignments may indicate potential model failures.  

    \item \label{t:anchor}
    \textit{Assess how segments jointly influence the generation.} 
    To gain a more precise understanding of the LLM's local behaviors, users assess how multiple segments together influence the model responses. Users select the important segments identified by the attributions or according to their prior knowledge and group all counterfactuals by these segments' occurrence. By inspecting each group, users can gain a precise understanding of these segments' influence on the model response. 

    \item \label{t:cf-explanation}
    \textit{Search counterfactual examples to validate findings.} 
    During the analysis, users also use concrete examples to help them gain a vivid understanding of the model's behavior. 
    From the examples, users can form new hypotheses about the segments' joint influence, e.g., a combination of some segments is sufficient to make consistent predictions and move back to the last task to test them.

\end{enumerate}

This workflow helps users align their mental model with the LLM's actual behavior, enabling them to assess its local performance on a given prompt and identify potential misalignments or failures.

\section{Algorithm Pipeline}

In this section, we present our approach to defining interpretable text representations and leveraging them to generate meaningful counterfactuals. 
We begin by introducing a novel computational pipeline that constructs interpretable representations of text and performs perturbations, i.e., generates counterfactuals based on these representations. 
We then describe an experiment designed to evaluate the effectiveness of the proposed algorithm in producing grammatically correct counterfactuals across five datasets from diverse domains.

\subsection{Define Interpretable Representations}

Creating interpretable representations of text can be framed as a text segmentation problem. 
The goal is to identify a segmentation that minimizes the number of segments while remaining expressive enough to encode all meaningful counterfactuals, i.e., all removal-based perturbations of this instance could be described as a subset of these segments. 

\smallskip

\begin{figure*}[htb]
    \centering
    \includegraphics[width=\textwidth]{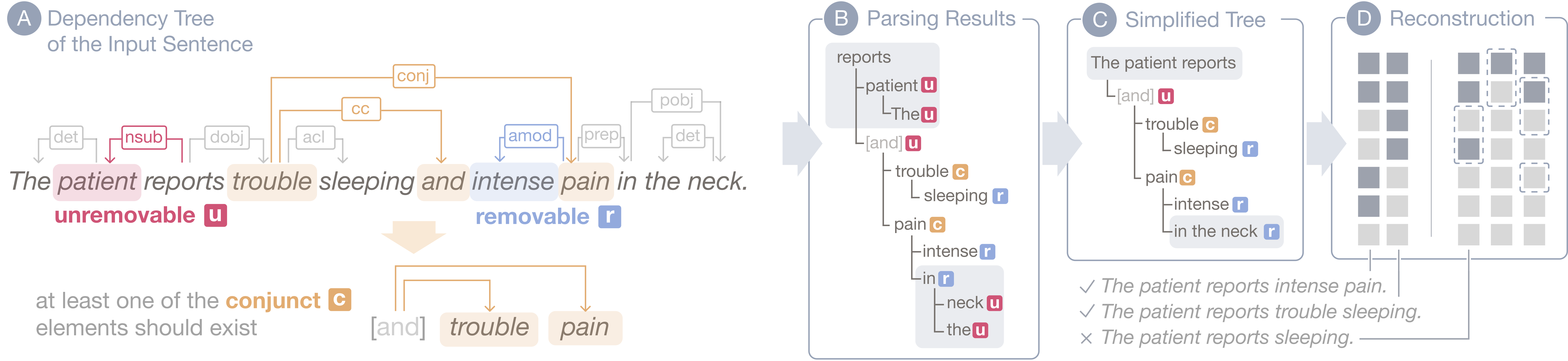}
    \caption{
    The algorithm takes a sentence as input, segments text into interpretable components, and outputs meaningful counterfactuals.
    The pipeline uses the sentence's dependency structure (A) to categorize the words as removable, unremovable, or conjunct (B). By grouping the connected unremovable words, we reduce the number of elementary segments (C). Finally, we filter complete sentences (i.e., counterfactuals) from all combinations according to the segments' correspondence relationship (D).
    }
    \label{fig:alg-pipeline}
\end{figure*}

To address this issue, we leverage the sentence’s dependency syntax to determine whether pairs of words should be grouped together.
For example, in the sentence ``\textit{The patient reports trouble sleeping and intense pain in the neck.}'', the word ``neck'' is the object of the preposition ``in'' (\texttt{pobj}; see \autoref{fig:alg-pipeline}A).
If the word ``in'' is removed, its dependent ``neck'' should also be removed to avoid producing an ungrammatical sentence. Conversely, if ``neck'' is removed while ``in'' remains, the preposition lacks an object, rendering the sentence meaningless. Therefore, these two words should be grouped together as a single interpretable component.

\smallskip

Building on this idea, we propose a computational pipeline that segments the input sentence into interpretable components, forming a simplified representation of the sentence as a binary vector (\autoref{fig:alg-pipeline}A–C). 
Based on this interpretable representation, the algorithm can systematically generate all meaningful removal-based counterfactuals (\autoref{fig:alg-pipeline}D). 
The detailed process is described below.

\smallskip

\smallskip

\par{\textit{Parsing}.}
The parsing stage aims to create a hierarchical representation of the text showing the dependency and removability of the words~(i.e., whether the sentence structure is broken by removing the words).
Identifying the dependency structure of the text is a long-lasting problem in syntax analysis. 
We built our algorithm on existing dependency parsing algorithms and implementations.
The next question we need to solve is identifying the removability of the words, which can be inferred from their dependency relation with the parent word.
For example, removing the nominal subject (\texttt{nsub}) breaks the structure, while removing an adjectival modifier (\texttt{amod}) usually doesn’t (\autoref{fig:alg-pipeline}A).
Using this idea, we divide the dependency relation into two categories according to linguistic grammar: relations that infer the words to be ``optional'' (i.e., removable) and the opposite (i.e., unremovable).
In the dependencies, conjunctions are exceptions
Conjunct components influence each other, i.e., removing only one will not break the sentence structure. 
So, for the conjunct (\texttt{conj}) relationship, we create a dummy token as the common parent of the two conjunct parts. 
We validated and iteratively refined the categorization rules by testing on samples from the MedQA dataset. 
Using these rules, the algorithm generates a parsing tree of the sentence (\autoref{fig:alg-pipeline}B). 

\smallskip

\par{\textit{Simplification}.} 
The parsing tree is sufficient to generate grammatically correct removal-only counterfactuals. 
However, a word-by-word partition is inefficient in computing counterfactuals and is unfriendly for users to understand. 
We noticed that this tree structure could be further simplified by combining unremovable segments (dummy segments excluded) with their parent segment (\autoref{fig:alg-pipeline}C), which reduces the number of segments.  
The outcome of this step identifies the elementary interpretable components of the sentence. 
In this example shown in \autoref{fig:alg-pipeline}C, the algorithm produces seven components (e.g., ``\textit{The patient reports}'', ``\textit{pain}'', and ``\textit{in the neck}''), including a dummy component, ``\textit{[and]}''. 
A meaningful removal-based perturbation on this sentence can then be represented by a binary vector $z' \in \{0,1\}^7$, where each dimension indicates whether the corresponding segment is included or removed (\autoref{fig:alg-pipeline}D). 

\subsection{Generate Meaningful Counterfactuals}
\label{sec:cf-gen}

Using the segments derived in the previous step, along with their correspondence relationships, we select meaningful counterfactuals from all possible combinations of these segments using three rules (\autoref{fig:alg-pipeline}D).
First, if the parent segment is removed, all segments in its branch should also be removed.
Second, if the parent segment is preserved, its unremovable child segments should also be preserved.
Third, if a dummy segment (e.g., \texttt{[and]}) is preserved, at least one of its children should also be preserved.
Finally, the algorithm reconstructed its corresponding textual counterfactuals for each valid combination.

\smallskip

\par{\textit{Customization on Demand}.}
The algorithm is extended to support users to customize the granularity of the segments by joining a non-leaf segment with all segments in its branch, for example, joining ``\textit{pain}'', ``\textit{intense}'', and ``\textit{in the neck}'' to a longer segment ``\textit{intense pain in the neck}''.
For each leaf segment, users can define alternative options for replacement, for example, replacing ``\textit{in the neck}'' with ``\textit{in the stomach}'' to generate diverging counterfactuals.

\subsection{Experiments}
\label{sec:experiment}

In this section, we present the experiments for evaluating the quality of the counterfactuals generated from the proposed pipeline.

\smallskip

\par{\textit{Metrics}.}
We evaluate the generated counterfactuals based on two criteria: diversity, measured by the number of unique counterfactuals, and fluency, assessed by their grammatical correctness.
A high diversity indicates the interpretable representation could describe a large population of counterfactual examples, i.e., being expressive.
And the fluency represents the quality of the counterfactuals, i.e., being interpretable to humans. 
We use \href{https://github.com/jxmorris12/language_tool_python}{\textit{LanguageTool}} to assess the grammatical correctness of the generated counterfactuals, which takes a sentence as input and outputs a list of grammatical and stylistic errors. 
Since the prototype sentences in the dataset have inherent grammatical errors, 
we measure if the generated counterfactuals introduce new grammatical errors.
If not, we count the generated counterfactual as grammatically correct. 

\smallskip

\par{\textit{Datasets}.}
We choose five datasets, including MedQA~\cite{madaan2021generate}, BillSum~\cite{kornilova2019billsum}, FiQA~\cite{maia2018financial}, TinyTextbooks~\cite{pham2023tiny}, and news--MultiNews~\cite{alex2019multinews}, that cover five common domains: medicine, legal, finance, education, and news reports. 
We randomly select 1,000 sentences from each dataset (5,000 in total) and generate counterfactuals by removal.

\smallskip

\par{\textit{Results}.}
The results are shown in \Cref{tab:parser_results} with an average parse and sample time per sentence of 0.7s using a laptop.
The most consequential contributor to this time is the dependency parsing by \href{https://spacy.io/models/en#en_core_web_trf}{Spacy's \texttt{trf}} Roberta Transformer model.
The grammaticality, a lower bound by the tool, is consistently above 95\% (average 97.2\%), and a single sentence usually generates $\sim$46 perturbations.
In \Cref{sec:perturbation_examples}, we show samples of the generated counterfactuals from all datasets.
We intended to use Polyjuice~\cite{wu2021polyjuice} (with the \texttt{delete} code), a general textual counterfactual example generation algorithm, as the baseline. 
However, in the original implementation, Polyjuice only outputs a few counterfactuals, which is not comparable with the proposed algorithm.
Running Polyjuice multiple times to get a comparable number of counterfactuals takes over an hour for each sentence, which is \re{not}{} adaptable in large-scale benchmarking. 
We make further discussions in \Cref{sec:polyjuice_why_not_good}.

\begin{table}[htbp]
\centering
\begin{tabular}{lccccc}
\toprule
\bf Dataset &
\bf Sent. length & 
\bf Pert./sent. & 
\bf Grammatical \\
\midrule
\href{https://huggingface.co/datasets/bigbio/med_qa}{MedQA} & 
13.3 & 51 & 98.3\% \\ %
\href{https://huggingface.co/datasets/billsum}{BillSum} & 
18.1 & 59 & 97.2\% \\ %
\href{https://huggingface.co/datasets/gbharti/finance-alpaca}{FiQA} &
16.2 & 46 & 95.9\% \\ %
\href{https://huggingface.co/datasets/nampdn-ai/tiny-textbooks}{TinyTextbooks} &
13.3 & 41 & 97.5\% \\ %
\href{https://huggingface.co/datasets/multi_news}{MultiNews} & 
15.6 & 34 & 97.1\% \\ %
\bottomrule
\end{tabular}
\caption{We evaluate the grammatical correctness of the generated counterfactuals using five datasets, covering five different domains for our pipeline. The average grammaticality rate is 97.2\%.}
\label{tab:parser_results}
\end{table}

\section{Interface Design}
\label{sec:interface}

We introduce \system{}, an interactive visualization tool for analyzing LLMs using counterfactuals. 
The system usage is divided into three stages: task creation, configuration, and results analysis. 

\subsection{Task Creation}
In the task creation stage, the users input the prototype text---the sentences to be perturbed and an optional prompt template. 
For example, for a prompt containing a multi-option question (the prompt text in \autoref{fig:teaser}C), the users don't want to perturb the question and options but only generate counterfactuals for the descriptions of the patient. 
They define the prompt template as ``\{input\} \textit{Which of the following is...}.''

\smallskip

In addition to providing input, users also need to define their hypotheses through evaluators. For instance, in the question-answering task described above, a user may want to know when the model outputs the correct answer (D. Nitrofurantoin). 
In this case, similar to prior work~\cite{arawjo2024chainforge}, our approach defines an evaluator as a single-rule binary classifier composed of an operator (e.g., \texttt{CONTAIN}) and the corresponding text (e.g., Nitrofurantoin), which together transform generated text into a Boolean outcome (e.g., including the word ``Nitrofurantoin'' or not).

\smallskip

We provide a set of predefined operators for users to specify the evaluators, including token-matching operators (e.g., \texttt{CONTAIN}, \texttt{STARTWITH}, and \texttt{EQUAL}) and logic-based operators (e.g., \texttt{ENTAIL}, \texttt{CONTRADICT}, and \texttt{SEMANTICEQUAL}). 
\re{Logic-based operations rely on entailment inference. We prompt the LLM to perform this calculation. 
In general, token-matching-based operators are used for concrete classification tasks, and logic-based operators should be used when the text generation task is open-ended. Typically, the user only needs to define an evaluator, for example, to determine whether the model makes the correct decision based on the original context. For more complex tasks, users can define multiple evaluators to handle different aspects of the generation.}{}
An example of the user interface is shown in \Cref{sec:user-study-design}.

\subsection{Experiment Panel}

\begin{figure}[!b]
    \centering
    \includegraphics[width=\linewidth]{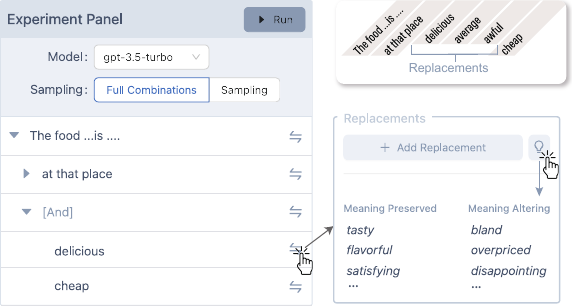}
    \caption{\re{The experiment panel allows users to customize counterfactual generation settings, such as adjusting segment granularity and specifying alternative words for replacement. The system generates two sets of candidate replacements to assist users in selecting suitable alternatives.}{}}
    \label{fig:panel-design}
\end{figure}

From the users' input text, the system uses the algorithm introduced in \Cref{sec:cf-gen} to divide text segments and present them in the experiment panel using a hierarchical list (\autoref{fig:panel-design}). 

\smallskip

Users can open and collapse text nodes using the experiment panel to change the granularities (\ref{t:create-exp}). 
Users can merge segments with their parent nodes to reduce complexity and expand relevant segments to explore finer-grained explanations.
\re{Additionally, users can input alternative text options for leaf segments to test the LLM’s behavior under different conditions. To streamline this process, the system leverages the LLM to suggest replacement candidates.
Specifically, the LLM generates two sets of potential replacements: one consisting of synonyms
or words that preserve the original meaning, and another containing alternatives that
significantly alter the meaning of the sentence. }{}

\smallskip

By clicking the ``Run'' button, the system will create counterfactuals based on users' configurations and get LLM responses.
\re{Most LLMs use probabilistic sampling when generating text, making their outputs inherently non-deterministic. To account for this variability, we query the model multiple times~\cite{cheng2024relic} ($n=5$) for each input in order to estimate the probability that the generated outputs satisfy the evaluator's criterion, which we refer to as the outcome.}{}

\subsection{Table View}

\begin{figure*}[!ht]
    \centering
    \includegraphics[width=\textwidth]{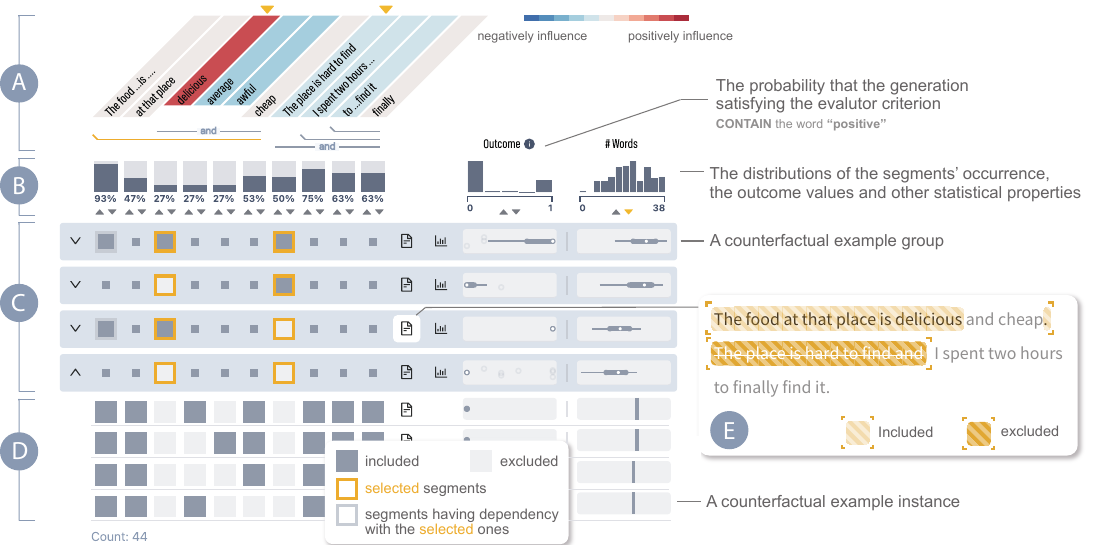}
    \caption{\system{} supports interactive analysis of counterfactuals through a table-based visualization. 
    (A) The header presents the text segments and their dependencies, where the color encodes the feature attributions. 
    (B) The bar charts and histograms visualize the distribution of the segments' occurrence, outcome values, and the number of words included in the counterfactuals. 
    Users can select segments and perform a \texttt{group-by} operation. 
    (C) \system{} then divides the counterfactual examples into groups and visualizes the aggregated properties in the table body. 
    (D) By expanding a group, users can inspect individual examples included in the group. 
    (E) For each group, \system{} also annotate the selected segments in the original text to help users understand the textual counterfactuals represented by the group.}
    \label{fig:interface}
\end{figure*}

\system{} uses a table-based visualization, inspired by UpSet~\cite{lex2014upset} and Juniper~\cite{nobre2018juniper}, to support interactive analysis of counterfactuals. 
The table view is composed of the following components.

\smallskip

\par{\textit{Segment Text and Attributions}.}
The header of the table presents the text segments with their attribution values (\autoref{fig:interface}A, \ref{t:attribution}). 
\re{Attribution values are computed using the KernelSHAP algorithm~\cite{lundberg2017shap}, which aggregates the outcomes of multiple perturbations to assign contributions to individual text segments. The algorithm details are introduced in \Cref{sec:kernelshap}. These attributions are then visualized using color encoding.}{We use color to encode the attribution values, calculated using KernalSHAP~\cite{lundberg2017shap}. 
The colors (red and blue) are friendly to colorblind people.}
\re{The red color indicates a positive influence, meaning that by keeping this segment, the LLM generation has a higher probability to satisfy the evaluator condition, i.e., having a higher outcome value.}{The red color indicates a positive influence, meaning that by keeping this segment, the LLM is more likely to have a positive prediction (i.e., having a higher outcome value).} 
On the contrary, segments in blue negatively influence the outcome---by removing the segment, the outcome value increases in general.

\smallskip

\par{\textit{Segment Dependencies}.} 
\re{
We annotate segment dependencies beneath the segment text (\autoref{fig:interface}A, bottom) to help users recognize structural relationships within the text. 
For instance, if a parent segment, such as the backbone of a sentence, is removed, its child segments should also be removed. 
Conjunction relationships are also indicated using underlines labeled with the conjunction type (e.g., and) at the center.
When a user selects a segment, all associated dependency underlines are highlighted to indicate the related segments that should be considered.
}{}

\smallskip

\par{\textit{Overview Charts}.}
When the table includes a large number of counterfactuals, 
users need to gain an overview understanding of the dataset~(\ref{t:explore-cfs}).
We use bar charts and histograms to visualize the distributions of the segment occurrence, outcome values, and the number of words~(\autoref{fig:interface}B) and support sorting and filtering interactions to find counterfactuals of interest (\ref{t:cf-explanation}).

\smallskip

\par{\textit{Counterfactual Examples and Groups}.}
\re{To present detailed counterfactual examples to users, we explored two design approaches. Our initial design displayed the perturbed texts in a list format. However, this approach did not scale well for longer texts, as it became difficult to navigate and compare. To address this issue, we adopted a table-based design that visualizes the compositional segments of the text, enabling more efficient comparison and interaction.}{}
The color in each cell represents whether the corresponding segment is included \furuiglyph{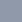} or not \furuiglyph{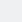} (\autoref{fig:interface}D). 
\re{Users can hover over or click the "Text" button to view the corresponding content.}{}
Each row also contains charts displaying the outcome value and the word count of the counterfactual example.
\system{} allows users to interactively aggregate the counterfactual examples by selecting segments and performing a \texttt{group-by} operation~(\ref{t:anchor}).
For example, in \autoref{fig:interface}, the user selected ``\textit{delicious}'' and ``\textit{The place is hard to find...}'' to understand the distributions of the outcome values under different combinations of their occurrence states.
Then, the system divides the instances into groups and visualizes their aggregated properties in the table body (\autoref{fig:interface}C).

\smallskip

We use boxplots to visualize the outcome distributions and highlight the selected segments in each group using squares with orange borders \furuiglyph{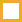}.
The occurrence of the selected segments may also influence other segments. For example, in groups where ``\textit{delicious}'' is always included, its parent segment ``\textit{The food is}...'' also exists in all counterfactual examples within the group. 
We highlight these influenced segments (due to the dependency) as well, using gray borders \furuiglyph{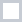}. 
Users can inspect the individual examples within the group by clicking the arrow button on the left. Users can filter these examples by brushing on the boxplot, e.g., to check outliers (\ref{t:cf-explanation}). 

\smallskip

\par{\textit{Text Annotations for Counterfactual Groups}.}
The graphical representation for counterfactual groups may not be intuitive to all users. So, we designed a text annotation schema that bordered and marked the included \furuiglyph{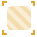} and excluded \furuiglyph{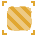} segments in the original input (\autoref{fig:interface}E).

\section{Evaluation}

We evaluate \system{} through a hypothetical use case, a user study, and feedback from XAI and NLP experts. 

\subsection{Hypothetical Use Case}
\label{sec:use-case}

We introduce a hypothetical scenario to demonstrate how the system supports the workflow and user tasks defined in \Cref{sec:requirements}.
In this scenario, a hypothetical user, Emma, was a trainee doctor.
She was interested in exploring new techniques for \re{automating}{automizing} clinical decisions and was specifically curious about how LLMs are capable of suggesting treatments. 
She selected a medical testing question about treatment selection for patients with urinary tract infections (UTIs) and fed it into GPT-3.5. 
While the model returned the correct answer (\texttt{D. Nitrofurantoin}), she wondered if the model took the proper information for making the prediction.
She input this task into our system and defined an evaluator as \texttt{CONTAIN(Nitrofurantoin)}.

\smallskip

\par{\textit{Customize the counterfactual generation (\ref{t:create-exp}).}}
The system suggested a list of text segments from the input prompt text, which are the basic elements for performing removal and replacements. 
After going through these segments, she found them all to be complete, reasonable phrases or sentence templates. 
Some of the segments included important information in this context, like ``\textit{pregnant}'' and ``\textit{at 22 weeks gestation}'', while others were less informative, like ``\textit{[She] otherwise [feels well]}''.
So, she combined these segments with their ancestor segments and reduced the number of segments from 26 to 14 (\autoref{fig:teaser}A). 
She kept other settings as default and submitted the configuration. 

\smallskip

\begin{figure}[tb]
    \centering
    \includegraphics[width=\linewidth]{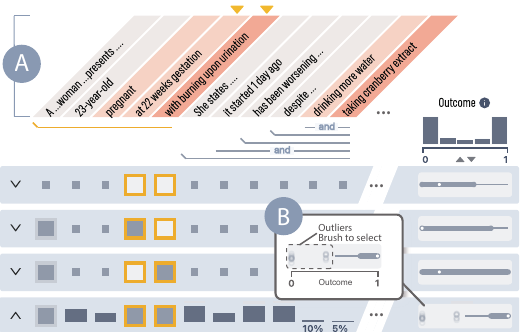}
    \caption{(A) The user identified the important segments (segments in red). To gain precise explanations, the user selected ``\textit{at 22 weeks gestation}'' and ``\textit{burning upon urination}'' based on her knowledge and grouped the counterfactuals. (B) In the group where both segments exist, most of the outcome values (probability) are high, indicating that these two segments are almost sufficient for the LLM to make consistent and correct predictions. However, there are outliers. Through brushing, the user selected these instances for further inspection.}
    \label{fig:case}
    \vspace{-1em}
\end{figure}

\par{\textit{Explore and Identify important segments (\ref{t:explore-cfs}, \ref{t:attribution}).}}
After the system returned the results, Emma first looked at the table header, where she noticed that ``\textit{burning upon urination}'' and another two segments were identified as having the highest positive contribution to the outcome (\autoref{fig:case}A). ``\textit{pregnant}'', ``\textit{at 22 weeks gestation}'', and the other two segments also have a noticeable positive contribution, and the last segment~(sentence) has a negative contribution.
In general, she agreed that the LLM identified the most critical information in the context---``\textit{burning upon urination}'' indicated the patient's symptom (e.g., UTI) and ``\textit{pregnant}'' suggested the doctor needed to carefully choose treatment where option \texttt{D} was the best choice. 
However, to take \texttt{Nitrofurantoin}, the patient should not be in the third trimester ($< 37$), which is guaranteed by ``\textit{at 22 weeks gestation}''.
To gain a more precise understanding, she interactively aggregated important segments to assess their anchoring effect. 

\smallskip

\par{\textit{Gain precise explanations (\ref{t:anchor}).}}
She grouped the counterfactual examples by ``\textit{at 22 weeks gestation}'' and ``\textit{burning upon urination}''. These two segments include sufficient information for her to make the prediction, and she wanted to know if the LLM's prediction is always aligned with hers.
She then looked at the group where the two segments were both contained. 
Through the outcome box plot (\autoref{fig:case}B), she noticed that most of the outputs had a high probability, suggesting that in most cases when the two segments and the sentence backbone (``\textit{A woman presents...}'') are included in the input prompt, the LLM would choose the correct answer, which aligned with her knowledge.

\smallskip

\par{\textit{Valid findings with examples (\ref{t:cf-explanation}).}}
However, there are exceptions. 
She found outliers in the box plot whose outcome probability was lower than 0.5. 
By brushing on the box plot, she selected these examples~(20 in total)~(\autoref{fig:teaser}D). 
These examples demonstrate that even if the input provides enough information (symptoms and pregnancy) for prediction, the model may still make an incorrect prediction, resulting in a failure.

\smallskip

\par
\begin{figure}[htb]
    \centering
    \includegraphics[width=0.8\linewidth]{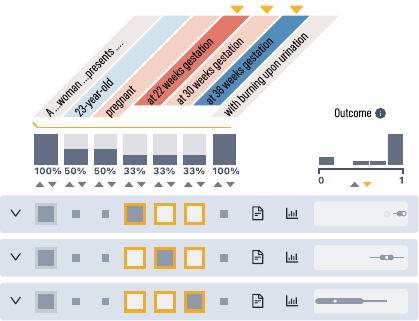}
    \caption{The user tested the LLM's treatment suggestion for patients with different gestation stages. She found that the model changed the suggested treatment when the patient was at 38 weeks gestation.}
    \label{fig:case-step2}
\end{figure}

{\textit{Change the settings for further analysis (\ref{t:create-exp}).}}
To understand if the model correctly captured the information in ``\textit{at 22 weeks gestation}'', the user conducted another round of analysis by replacing the gestation stage to ``\textit{30 weeks}'' and ``\textit{38 weeks}''.
If the LLM used this information correctly, after replacing it with ``\textit{38 weeks}'', the LLM will give an alternative prediction (i.e., recommending a different treatment) since option \texttt{D} should not be given to patients in the third trimester ($\geq 37$).
For simplicity, she excluded other sentences. Then, she submitted the configuration.
From the results (\autoref{fig:case-step2}), she noticed that when replaced by ``\textit{at 38 weeks gestation}'', the corresponding outcome values are in general negative ($< 0.5$), indicating that the LLM made use of this information.
However, not all outcome values are lower than 0.5, which suggests that the LLM may still recommend \texttt{Nitrofurantoin} to pregnant patients in the third trimester, which is risky. 

\smallskip

\par{\textit{Conclusions.}}
From the above analysis, Emma identified the segments with higher contributions to the predictions. 
Through interactive aggregation and inspecting concrete examples, she gained a vivid understanding of how these segments contribute to the predictions. 
She also found misalignments with the LLM in suggesting treatment for UTI patients, where the LLM may give unexpected predictions when ``\textit{despite taking cranberry extract}'' is not explicitly mentioned in the text or for the patients at 38 weeks gestation.
These findings suggest that the LLM should be used cautiously in such treatment suggestion tasks.

\subsection{User Study}

\re{
We conducted a user study with eight participants (P1–P8) to explore how users interact with the system when interpreting LLM behaviors and to evaluate its usability and usefulness.
Participants (ages 23–32, average: 26.6) were recruited through a university mailing list.
All participants had prior experience with LLMs.
Seven out of eight participants were familiar with explainable XAI methods, with feature attribution being the most commonly used approach among them.
}{We conducted a user study to assess the efficacy, usability, and usefulness of \system{} in helping users understand LLM behaviors.}

\smallskip

\par{\textit{Procedure.}}
The user study lasted approximately 60 minutes.
At the beginning, we introduced the background of the study to the participants, obtained their consent, and collected demographic information.
We then demonstrated the use of \system{} through a sentiment classification example. Participants were encouraged to freely explore the system using this example and to ask questions as needed.
Once they were familiar with the system, we presented them with an instance from HotpotQA~\cite{yang2018hotpotqa} along with GPT-3.5's prediction for that instance. We introduce the details of this task in \Cref{sec:user-study-design}.
Participants explored model behaviors through \textit{What if} scenarios, answering three key questions (\textit{Why (not)}, \textit{How to be that}, \textit{How to still be this}) within 20 minutes while thinking aloud.
After the task, we conducted semi-structured interviews to gather feedback on their experience, including helpful features, challenges, real-world applications, and suggestions.
Participants also completed a System Usability Scale questionnaire (5-point Likert scale)~\cite{brooke1996susa}.
The entire session was audio- and screen-recorded, and transcripts were analyzed to extract key insights presented below.

\subsubsection{Usability and Usefulness}

\begin{figure}[t]
\centering
\includegraphics[width=\linewidth,trim={0 11mm 0 0},clip]{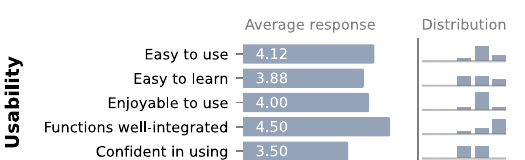}

\vspace{1mm}
\includegraphics[width=\linewidth,trim={0 0 0 6mm},clip]{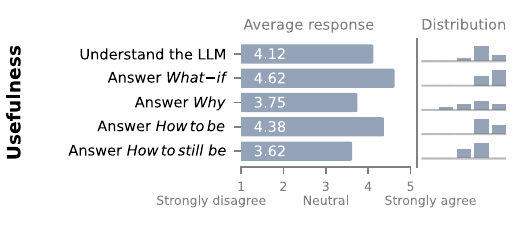}

\caption{\re{Results from the usability and usefulness questionnaire for \system{}. The eight participants reported high usability and usefulness.}{}}
\label{fig:questionnaire}
\vspace{-1em}
\end{figure}

\par{\re{\textit{Overall Usability.}}{}}
\re{In general, the participants in our study found \system{} easy to use~($\bar{x} = 4.12$; see \autoref{fig:questionnaire}).
P2 noted that “\textit{the color scale is relatively easy to understand},” a sentiment echoed by P4, who added, “\textit{it really gives already an intuitive overview of what to look for}.”
Almost all participants (P1-4, P6, and P8) also emphasized the value of the \texttt{group-by} operation. 
P3 commented that it allows to ``\textit{control [the presentation] directly}'' and reduce the information presented.
P8 suggested that the \texttt{group-by} operation allows for a deeper exploration of the data.
P1 concluded that ``\textit{the combination [of the system features] is always important}'' and \system{} integrates these features well, a view shared by all other participants~($\bar{x} = 4.5$).}{}

\smallskip

\par{\re{\textit{Learning Barriers.}}{}}
\re{
Overall, participants agreed that the system is easy to learn~($\bar{x} = 3.88$) and enjoyable to use~($\bar{x} = 4.00$).
However, they also noted that some functionalities were not immediately intuitive.
Both P3 and P7 mentioned that they found the sentence dependencies to be hard to read.
P2 expressed feeling overwhelmed by “\textit{the matrix layout},” though they acknowledged that “\textit{with the grouping, it is a bit better}.”
P8 initially found the box plots difficult to read, but gained understanding after receiving an explanation of how the visualization works.
This complexity influenced the user confidence, as reflected in the confidence rating~($\bar{x} = 3.50$).}{}

\smallskip

\par{\textit{\re{Perceived Usefulness.}{}}}
\re{Participants generally found \system{} helpful for understanding LLM behaviors ($\bar{x} = 4.12$), particularly when exploring \textit{what-if} scenarios ($\bar{x} = 4.62$) and addressing \textit{How to be} questions ($\bar{x} = 4.38$) through the system’s intuitive sorting and filtering interactions. In contrast, \textit{Why} ($\bar{x} = 3.75$) and \textit{How to still be} ($\bar{x} = 3.62$) questions were perceived as more challenging. For example, P5 and P7 reported difficulty interpreting SHAP values, while P3 noted that answering \textit{How to still be} questions was complicated by exploring the large solution space, which made it harder to reach a clear conclusion.
Despite these challenges, participants consistently affirmed the value of \system{} in supporting LLM behavior understanding. All participants emphasized its usefulness in assessing the trustworthiness of the model's behavior in relation to the problem at hand.}{}

\subsubsection{Usage Patterns}

\par{\textit{Viewing feature attributions as the first step.}} 
We observed that all participants began by viewing the headers, which use color to encode the attribution of each segment.
All participants reported that this feature helps them answer the \textit{why} question by identifying the most influential segments.
P1 also noted that, by identifying the important segments, they could hypothesize whether a subset of these segments was sufficient to ``anchor'' the prediction.
They then grouped the results based on the presence of these segments to explore different combinations, which helped them find the answer to the \textit{how to still be} question.

\smallskip
\par{\re{\textit{Frequently using filtering to identify useful examples.}}{}}
\re{Filtering emerged as the most frequently used interaction. We observed two primary usage patterns of the filtering operation.
The first pattern involved filtering instances with differing outcomes and sorting them by the number of remaining words to identify minimal perturbations that alter the model’s prediction—a strategy adopted by nearly all participants. As demonstrated in the example session, participants quickly learned and effectively applied this method during the task. 
For instance, P3 discovered that removing two seemingly irrelevant segments caused the model to produce an incorrect prediction and found a lack of robustness in the model’s reasoning.
The second pattern involved hypothesis testing through selective filtering. Participants filtered for instances containing information they believed to be either sufficient for a correct prediction or lacking necessary context (P1-3, P5, P8). This approach allowed them to actively probe and uncover model flaws by validating or refuting their assumptions about what the model relied on to make decisions.}
{}

\smallskip
\par{\textit{Checking corresponding text to confirm findings.}} 
We observed that users frequently referred to the corresponding text of counterfactuals or counterfactual groups (e.g., \Cref{fig:interface}E).
Although the matrix-based visualization is more space-efficient for representing a population of counterfactuals, participants still relied on reading the actual text to gain a clearer and more intuitive understanding of what each counterfactual or counterfactual group is about. \re{In this spirit, P7 commented ``\textit{seeing the actual text, ..., seeing which words you removed or not is essential.}"}{}

\subsection{Expert Interview}

\re{We further conducted interviews with three NLP researchers (E1-3) and three XAI researchers (E4–E6). Each has between three and over ten years of experience publishing in the field. All are either LLM practitioners themselves or work closely with LLM practitioners, enabling them to offer insights from a practitioner’s perspective that complements the viewpoints captured in the earlier user study.}{}

\smallskip
\par{\textit{Procedure.}}
We conducted semi-structured interviews with the participants. 
Each interview was composed of three sessions and took around 45 minutes. 
In the first session, we introduced the background and demonstrated our system using the use case in \Cref{sec:use-case}.
Then, we allowed them to use our system for free explorations. 
In the final session, we collected their feedback about their overall impression of the system, potential usage scenarios, and desired improvements.

\smallskip
\par{\textit{Overall Assessment.}}
Experts from both domains found the system design to be concise and easy to comprehend, presenting a comprehensive overview of numerous functionalities.
E3 highlighted that our way of the feature attributions is the most intuitive explanation to lay users. 
E4 and E6 highly praised the way we integrated multiple explanation techniques. 
``\textit{Attributions are not sufficient for users to make decisions; the concrete (counterfactual) examples help.}'' suggested by E6.

\smallskip

\par{\textit{Multi-Level Segmentations.}}
Most experts highlighted the positive implications of our multi-level segmentation for language model explainability. E5 described how traditional XAI methods are not developed for NLP and tend to underperform when applied without careful consideration of syntactical constraints: ``\textit{They were not created for text data. In my opinion, we should consider the relationship and interaction between words.}'' 
Similarly, E6 discussed that creating meaningful perturbations for attribution methods, such as LIME\cite{ribeiro2018lime}, remains a significant challenge in NLP. They agreed that our way of sampling perturbations based on the dependency structure provides a helpful way to address this issue: ``\textit{you take LIME and improve sampling, which is the most critical part, I think, of the algorithm}.''

\smallskip
\par{\textit{Adaptions to Different NLP Tasks.}}
Throughout our interviews, the NLP researchers also highlighted potential usage scenarios of the proposed system in more challenging tasks.
E2 and E5 were interested in using the system to solve more open-ended language generation tasks, such as text summarization. We further discuss it in \Cref{sec:discussion}.
Besides, E1 and E3 were interested in using the system to explain model reasoning through attribution in math world problems and as a potentially useful tool for causal mediation analysis.
Finally, E2 suggested a potential extension of supporting multi-round dialog input.

\section{Discussion}
\label{sec:discussion}

Based on feedback from both users and experts, we outline two directions for future research and discuss the limitations of our system.

\smallskip
\par{\re{\textit{Understanding the Dynamics of the Explanation Process.}}{}}
\re{The explanation process is inherently interactive, evolving through a dynamic exchange between the explainer and the explainee~\cite{miller2019explanation}.
A deeper understanding of this process can inform the development of more effective explanation tools that enhance both explanatory efficiency and user experience. In this paper, we introduce a workflow for explanation in \Cref{sec:requirements} and design \system{} to support this workflow in practice. Looking ahead, this system can serve as a design probe to investigate how users engage with different types of explanations when solving real-world problems, enabling empirical insights into explanation usage patterns and informing future system design.
}{}

\smallskip
\par{\re{\textit{Adapting to Free-Form Text Generations.}}{}}
\re{The system is designed to help users analyze and explain LLMs. 
It requires the users to define concrete evaluators---they need to specify what they want to explain. 
This is relatively straightforward when the model's task is inherently a classification problem, such as multiple-choice question answering or sentiment analysis. However, for free-form text generation tasks, such as creative writing or complex reasoning, defining appropriate evaluators becomes much more challenging.
We believe that logic-based evaluators (e.g., ENTAIL) offer a promising solution in these cases, as they are more general and better equipped to handle the variability inherent in natural language. 
Additionally, the system should support users in iteratively refining their evaluators based on the outputs. Enabling such exploratory analysis for open-ended text generation remains an important and open research challenge.}{The system is designed to help users analyze and explain LLMs. 
It requires the users to have concrete evaluators---knowing what to explain. 
The evaluators are easy to define for classification tasks like multi-option question-answering and sentiment prediction. 
However, the evaluators are not apparent for free-form text generation, like prompting the LLM with ``\textit{Given the above context, why should the patient be given this treatment?}''. 
To support free-form text generation tasks, the system should allow users to explore the text generations from different prompts and iteratively refine the evaluators. Supporting exploratory analysis on free-form text generation is an interesting open problem that requires further research.
}

\smallskip
\par{\textit{Limitations.}}
Our system enhances the flexibility of generating explanations compared to static explanations. However, this increased flexibility can also introduce additional cognitive workload, as users may need to make more decisions or navigate more options.
To strike a balance between flexibility and usability, we recommend strengthening the system’s guidance features, such as through intelligent suggestions or adaptive interface support. 
\re{
In addition, our current evaluation is limited to a small group of users within experimental settings. These evaluations focus on how well the system supports typical explanatory questions, but they are not grounded in concrete application scenarios such as model debugging or decision-making. Moving forward, we aim to assess the system’s real-world applicability by conducting evaluations with a larger and more diverse user base, spanning a broader range of practical use cases and deployment contexts.
}{For the study, our evaluation is limited to a small group of users within experimental environments. Moving forward, we plan to expand our evaluation to include a larger and more diverse user base across a broader range of real-world scenarios.}

\section{Related Work}
We discuss the connections between the proposed visualization and the counterfactual generation algorithm with existing approaches. 

\subsection{Visualizations for Understanding Language Model}

Visualizations play an essential role in explaining language models~\cite{brath2023role}. 
Existing methods can be categorized by their generalizability into model-agnostic and model-specific approaches. 

\smallskip

Most existing language-model-oriented visualization tools target supporting professional users (e.g., ML developers) to look into the model's inner structure for understanding and debugging~\cite{ming2017understanding, strobelt2017lstmvis, strobelt2019seq2seqvis,  tenney2020language, wang2021dodrio, spinner2024generaitor, shao2023visual}. %
RNNVis~\cite{ming2017understanding} uses a bipartite graph to help users understand the function of hidden state units in a recurrent neural network. LSTMVis~\cite{strobelt2017lstmvis} visualizes the dynamics of the neuron activations in an LSTM model to help users find activation patterns for analysis and debugging.
Various existing approaches visualize the attention layers of transformer models for explanations~\cite{wang2021dodrio, strobelt2019seq2seqvis, tenney2020language, yeh2023attentionviz}.
Seq2seq-Vis \cite{strobelt2019seq2seqvis} visualizes the attention and other modules within a sequence-to-sequence model to help users identify the failing part. 
To apply these model-specific approaches, we usually need to probe the language model's layer-wise outputs, which is not feasible in today's commercial models like GPT-4. 
Besides, when the language model's number of parameters and layers increases, understanding its inner mechanism becomes challenging.

\smallskip

Model-agnostic approaches, on the other hand, are more generally applicable. Such approaches usually focus on probing model behavior using input variations \cite{wexler2020whatif, coscia2023knowledgevis}. 
Existing work uses feature attribution methods to quantify each word's contribution to prediction and visualize them using a heatmap~\cite{wang2024commonsensevis, coscia2024iscore}.  
However, most of the model-agnostic feature attribution methods, especially the Shapley-based methods~\cite{lundberg2017shap}, cannot be scaled to support long-form inputs due to their exponential time complexity. 
Other model-agnostic approaches focus on understanding LLM's performance on the whole dataset~\cite{kahng2024llm, desmond2024evalullm, lee2025towards}. 
For example, LLM Comparator \cite{kahng2024llm} allows users to evaluate and compare the performances of two LLMs on a dataset using user-defined criteria. 

\smallskip

Our system introduces a model-agnostic framework for analyzing large language model (LLM) behavior at the instance level. Unlike existing local explanation methods, it scales more effectively to long-form text by employing a structured and efficient representation of the input. Additionally, \system{} combines aggregated feature attribution scores with representative examples in an intuitive visualization, enabling dynamic and interpretable analysis of model behavior.

\subsection{Creating Textual Counterfactuals}

Counterfactuals are an essential concept in causal reasoning and analysis, which means hypothetical ``what-if'' conditions~\cite{pearl2009causal, lewis2013counterfactuals, morgan2015counterfactuals, wang2024empirical}.
People use counterfactuals to assess causations by mentally simulating the outcomes under counterfactual conditions and comparing them with the factual results, known as counterfactual assessments or counterfactual reasoning~\cite{kahneman1981simulation}.
In NLP, a counterfactual is generally defined as a meaningful variation or perturbation from the prototype sentences~\cite{wu2021polyjuice}, which are used in language model assessments and explanations.

\smallskip

In model evaluation, researchers and developers often generate counterfactuals to assess specific aspects of model behavior—for example, testing whether the model produces consistent outputs for semantically equivalent inputs~\cite{ribeiro2020beyond}, or investigating hypotheses such as whether gender-related information significantly influences predictions.
In some studies, these counterfactual datasets are created manually~\cite{kaushik2020learning}, e.g., through crowd-sourcing, which is time-consuming. 
Other methods automate this process by using templates~\cite{ribeiro2020beyond} or language models to generate meaningful perturbed text~\cite{madaan2021generate, robeer2021generating}. 
For instance, Polyjuice~\cite{wu2021polyjuice} employs a fine-tuned GPT-2 model to enable controllable counterfactual generation. 
It allows users to create common types of counterfactuals, such as through negation, entity shuffling, deletion, or insertion. 
However, using these methods requires that developers already have specific hypotheses in mind-they must decide which property of the text they intend to manipulate, which has a limited usage in supporting explanation scenarios where users don't have such hypotheses.  

\smallskip

In explanatory settings, counterfactual explanations, which describe the minimal perturbations to the model input required to alter the model prediction~\cite{wachter2017counterfactual} are commonly used to explain a model prediction~\cite{wexler2020whatif} and probe the model's decision boundaries~\cite{cheng2020dece}. 
Removal-based methods such as LIME~\cite{ribeiro2018lime} and SHAP~\cite{lundberg2017shap} also commonly generate variations of the input text to probe the model’s local behavior. 
However, these approaches typically present only aggregated results (i.e., feature attributions) while obscuring the underlying sampling process and the model’s predictions for individual variations, even though the concrete examples may also convey model insights. 
This is largely due to the inefficiency of exploring large sets of examples without proper structure or organization. Besides, these examples commonly include meaningless sentences, hindering users' interpretations.

\smallskip

In this work, we create meaningful counterfactuals to answer the four types of explanatory questions presented in \Cref{sec:questions}.
It requires the counterfactuals to be meaningful (different from existing removal-based methods) and hypothesis-free (different from assessment-driven methods).
The most relevant approach is Polyjuice's deletion generation~\cite{wu2021polyjuice}. 
Through the experiment in ~\Cref{sec:experiment}, we demonstrate the superiority of our algorithm in terms of time efficiency and the fluency and diversity of the generated counterfactuals.

\smallskip

\section{Conclusion}

This paper presents \system{}, an interactive visualization system with an efficient counterfactual generation algorithm designed to support LLM practitioners and users in understanding LLM behaviors. The system facilitates interactive counterfactual generation and analysis, enabling users to actively engage in the exploration of LLM responses by varying target instances and analyzing outcomes at customizable levels of granularity.
We conducted experiments demonstrating that our counterfactual generation algorithm is both time-efficient and capable of producing high-quality counterfactuals. Additionally, a user study and expert interviews with professionals in NLP and XAI validate the system’s usability and usefulness.
Our findings underscore the importance of involving humans as active participants in the explanation process, rather than as passive recipients of explanations.

\bibliographystyle{misc/abbrv-doi-hyperref}

\bibliography{bibliography.bib}

\clearpage
\appendix

\section{Perturbation Examples}
\label{sec:perturbation_examples}

We intentionally selected shorter sentences in the interest of space and each output is thus on a separate line.
Sentences which are likely incorrect are marked by $\dagger$.
The outputs for Polyjuice are sampled once for each example with 100 prompts.

\textexamplenofig{
\textbf{A 23-year-old pregnant woman at 22 weeks gestation presents with burning upon urination.} (MedQA) \\[0.6em]

\hrule \vspace{0.6em}

\textbf{Ours:} \\
A woman presents.  \\
A pregnant woman presents.  \\
A 23-year-old woman presents.  \\
A woman presents with burning.  \\
A 23-year-old pregnant woman presents.  \\
A pregnant woman presents with burning.  \\
A woman at 22 weeks gestation presents.  \\
A woman presents with burning upon urination.  \\
A 23-year-old woman presents with burning.  \\
A pregnant woman at 22 weeks gestation presents.  \\
A pregnant woman presents with burning upon urination.  \\
A 23-year-old pregnant woman presents with burning.  \\
A 23-year-old woman at 22 weeks gestation presents.  \\
A 23-year-old woman presents with burning upon urination.  \\
A woman at 22 weeks gestation presents with burning.  \\
A 23-year-old pregnant woman at 22 weeks gestation presents.  \\
A 23-year-old pregnant woman presents with burning upon urination.  \\
A pregnant woman at 22 weeks gestation presents with burning.  \\
A woman at 22 weeks gestation presents with burning upon urination.  \\
A 23-year-old woman at 22 weeks gestation presents with burning.  \\
A pregnant woman at 22 weeks gestation presents with burning upon urination.  \\
A 23-year-old pregnant woman at 22 weeks gestation presents with burning.  \\
A 23-year-old woman at 22 weeks gestation presents with burning upon urination.  \\
A 23-year-old pregnant woman at 22 weeks gestation presents with burning upon urination.  \\[0.6em]

\hrule \vspace{0.6em}

\textbf{Polyjuice:} \\
A 23-year pregnant woman at 22 weeks gestation presents with burning upon urination.  \\
A 23-year-old pregnant woman at 22 weeks gestation presents with burning upon.  \\
22 weeks gestation presents with burning upon urination.  \\
pregnant woman $\dagger$ \\
A 23-year-old pregnant woman at 22 gestation presents with burning upon urination.
}

\textexamplenofig{
\textbf{A mother brings her 3-week-old infant to the pediatrician’s office because she is concerned about his feeding habits.} (MedQA)  \\[0.6em]

\hrule \vspace{0.6em}

\textbf{Ours:} \\
A mother brings her infant.  \\
A mother brings her 3-week-old infant.  \\
A mother brings her infant because she is concerned.  \\
A mother brings her infant to the pediatrician’s office.  \\
A mother brings her 3-week-old infant because she is concerned.  \\
A mother brings her infant because she is concerned about his feeding habits.  \\
A mother brings her 3-week-old infant to the pediatrician’s office.  \\
A mother brings her infant to the pediatrician’s office because she is concerned.  \\
A mother brings her 3-week-old infant because she is concerned about his feeding habits.  \\
A mother brings her 3-week-old infant to the pediatrician’s office because she is concerned.  \\
A mother brings her infant to the pediatrician’s office because she is concerned about his feeding habits.  \\
A mother brings her 3-week-old infant to the pediatrician’s office because she is concerned about his feeding habits.  \\[0.6em]

\hrule \vspace{0.6em}

\textbf{Polyjuice:} \\
mother brings her 3-week-old infant to the pediatrician’s office because she is concerned about his feeding habits. $\dagger$ \\
A mother brings her 3-week infant to the pediatrician’s office because she is concerned about his feeding habits.  \\
A mother brings her 3 infant to the pediatrician’s office because she is concerned about his feeding habits.  \\
A mother brings her 3-old infant to the pediatrician’s office because she is concerned about his feeding habits.  \\
A mother brings her 3-week-old infant to because she is concerned about his feeding habits.  \\
A mother brings her 3-week-old infant to the because she is concerned about his feeding habits.  \\
A mother brings her 3-week-old infant to the pediatrician because she is concerned about his feeding habits.  \\
A mother brings her 3-week-old to the pediatrician’s office because she is concerned about his feeding habits.  \\
A mother brings her 3-week-old to the  \\
}

\textexamplenofig{
\textbf{Requires States to allocate funds from Federal and State shares of program costs to LEAs according to specified formulae.} (BillSum) \\[0.6em]

\hrule \vspace{0.6em}

\textbf{Ours:} \\
Requires States to allocate funds.  \\
Requires States to allocate funds to LEAs.  \\
Requires States to allocate funds according.  \\
Requires States to allocate funds to LEAs according.  \\
Requires States to allocate funds from Federal shares.  \\
Requires States to allocate funds from State shares.  \\
Requires States to allocate funds according to formulae.  \\
Requires States to allocate funds from Federal and State shares.  \\
Requires States to allocate funds from Federal shares to LEAs.  \\
Requires States to allocate funds from Federal shares according.  \\
Requires States to allocate funds from State shares to LEAs.  \\
Requires States to allocate funds from State shares according.  \\
Requires States to allocate funds to LEAs according to formulae.  \\
Requires States to allocate funds according to specified formulae.  \\
Requires States to allocate funds from Federal and State shares to LEAs.  \\
Requires States to allocate funds from Federal and State shares according.  \\
Requires States to allocate funds from Federal shares of program costs.  \\
Requires States to allocate funds from Federal shares to LEAs according.  \\
Requires States to allocate funds from State shares of program costs.  \\
Requires States to allocate funds from State shares to LEAs according.  \\
Requires States to allocate funds to LEAs according to specified formulae.  \\
Requires States to allocate funds from Federal and State shares of program costs.  \\
Requires States to allocate funds from Federal and State shares to LEAs according.  \\
Requires States to allocate funds from Federal shares of program costs to LEAs.  \\
Requires States to allocate funds from Federal shares of program costs according.  \\
Requires States to allocate funds from Federal shares according to formulae.  \\
Requires States to allocate funds from State shares of program costs to LEAs.  \\
Requires States to allocate funds from State shares of program costs according.  \\
Requires States to allocate funds from State shares according to formulae.  \\
Requires States to allocate funds from Federal and State shares of program costs to LEAs.  \\
Requires States to allocate funds from Federal and State shares of program costs according.  \\
Requires States to allocate funds from Federal and State shares according to formulae.  \\
Requires States to allocate funds from Federal shares of program costs to LEAs according.  \\
Requires States to allocate funds from Federal shares to LEAs according to formulae.  \\
Requires States to allocate funds from Federal shares according to specified formulae.  \\
Requires States to allocate funds from State shares of program costs to LEAs according.  \\
Requires States to allocate funds from State shares to LEAs according to formulae.  \\
Requires States to allocate funds from State shares according to specified formulae.  \\
Requires States to allocate funds from Federal and State shares of program costs to LEAs according.  \\
Requires States to allocate funds from Federal and State shares to LEAs according to formulae.  \\
Requires States to allocate funds from Federal and State shares according to specified formulae.  \\
Requires States to allocate funds from Federal shares of program costs according to formulae.  \\
Requires States to allocate funds from Federal shares to LEAs according to specified formulae.  \\
Requires States to allocate funds from State shares of program costs according to formulae.  \\
Requires States to allocate funds from State shares to LEAs according to specified formulae.  \\
Requires States to allocate funds from Federal and State shares of program costs according to formulae.  \\
Requires States to allocate funds from Federal and State shares to LEAs according to specified formulae.  \\
Requires States to allocate funds from Federal shares of program costs to LEAs according to formulae.  \\
Requires States to allocate funds from Federal shares of program costs according to specified formulae.  \\
Requires States to allocate funds from State shares of program costs to LEAs according to formulae.  \\
Requires States to allocate funds from State shares of program costs according to specified formulae.  \\
Requires States to allocate funds from Federal and State shares of program costs to LEAs according to formulae.  \\
Requires States to allocate funds from Federal and State shares of program costs according to specified formulae.  \\
Requires States to allocate funds from Federal shares of program costs to LEAs according to specified formulae.  \\
Requires States to allocate funds from State shares of program costs to LEAs according to specified formulae.  \\
Requires States to allocate funds from Federal and State shares of program costs to LEAs according to specified formulae.   \\[0.6em]

\hrule \vspace{0.6em}

\textbf{Polyjuice:} \\
Requires States to allocate funds to LEAs according to specified formulae.  \\
Requires States to allocate funds from Federal and State shares of program to LEAs according to specified formulae.  \\
Requires States to allocate funds from Federal and State shares of program costs to LEAs according to.  \\
}

\textexamplenofig{
\textbf{Requires plaintiffs who obtain a preliminary injunction or administrative stay in Indian energy related actions to post bond.} (BillSum)  \\[0.6em]

\hrule \vspace{0.6em}

\textbf{Ours:} \\
Requires plaintiffs to post bond.  \\
Requires plaintiffs who obtain an injunction to post bond.  \\
Requires plaintiffs who obtain stay to post bond.  \\
Requires plaintiffs who obtain a preliminary injunction to post bond.  \\
Requires plaintiffs who obtain an injunction or stay to post bond.  \\
Requires plaintiffs who obtain administrative stay to post bond.  \\
Requires plaintiffs who obtain a preliminary injunction or stay to post bond.  \\
Requires plaintiffs who obtain an injunction in actions to post bond.  \\
Requires plaintiffs who obtain an injunction or administrative stay to post bond.  \\
Requires plaintiffs who obtain a preliminary injunction in actions to post bond.  \\
Requires plaintiffs who obtain a preliminary injunction or administrative stay to post bond.  \\
Requires plaintiffs who obtain an injunction in Indian actions to post bond.  \\
Requires plaintiffs who obtain an injunction in energy related actions to post bond.  \\
Requires plaintiffs who obtain an injunction in actions or stay to post bond.  \\
Requires plaintiffs who obtain a preliminary injunction in Indian actions to post bond.  \\
Requires plaintiffs who obtain a preliminary injunction in energy related actions to post bond.  \\
Requires plaintiffs who obtain a preliminary injunction in actions or stay to post bond.  \\
Requires plaintiffs who obtain an injunction in Indian energy related actions to post bond.  \\
Requires plaintiffs who obtain an injunction in Indian actions or stay to post bond.  \\
Requires plaintiffs who obtain an injunction in energy related actions or stay to post bond.  \\
Requires plaintiffs who obtain an injunction in actions or administrative stay to post bond.  \\
Requires plaintiffs who obtain a preliminary injunction in Indian energy related actions to post bond.  \\
Requires plaintiffs who obtain a preliminary injunction in Indian actions or stay to post bond.  \\
Requires plaintiffs who obtain a preliminary injunction in energy related actions or stay to post bond.  \\
Requires plaintiffs who obtain a preliminary injunction in actions or administrative stay to post bond.  \\
Requires plaintiffs who obtain an injunction in Indian energy related actions or stay to post bond.  \\
Requires plaintiffs who obtain an injunction in Indian actions or administrative stay to post bond.  \\
Requires plaintiffs who obtain an injunction in energy related actions or administrative stay to post bond.  \\
Requires plaintiffs who obtain a preliminary injunction in Indian energy related actions or stay to post bond.  \\
Requires plaintiffs who obtain a preliminary injunction in Indian actions or administrative stay to post bond.  \\
Requires plaintiffs who obtain a preliminary injunction in energy related actions or administrative stay to post bond.  \\
Requires plaintiffs who obtain an injunction in Indian energy related actions or administrative stay to post bond.  \\
Requires plaintiffs who obtain a preliminary injunction in Indian energy related actions or administrative stay to post bond.  \\[0.6em]

\hrule \vspace{0.6em}

\textbf{Polyjuice:} \\
Requires plaintiffs who obtain a preliminary injunction or administrative stay in Indian energy related actions to.  \\
Requires plaintiffs who obtain a preliminary injunction or stay in Indian energy related actions to post bond.  \\
Requires who obtain a preliminary injunction or administrative stay in Indian energy related actions to post bond.  \\
Requires plaintiffs who obtain a preliminary injunction or administrative stay to post bond.  \\
Requires preliminary injunction or administrative stay before it may be served.. $\dagger$ \\
}

\textexamplenofig{
\textbf{He didn’t take responsibility for his comment and he fails horribly at humor.} (MultiNews) \\[0.6em]

\hrule \vspace{0.6em}

\textbf{Ours:} \\
He fails.  \\
He didn’t take responsibility.  \\
He fails horribly.  \\
He fails at humor.  \\
He didn’t take responsibility and he fails.  \\
He fails horribly at humor.  \\
He didn’t take responsibility for his comment.  \\
He didn’t take responsibility and he fails horribly.  \\
He didn’t take responsibility and he fails at humor.  \\
He didn’t take responsibility for his comment and he fails.  \\
He didn’t take responsibility and he fails horribly at humor.  \\
He didn’t take responsibility for his comment and he fails horribly.  \\
He didn’t take responsibility for his comment and he fails at humor.  \\
He didn’t take responsibility for his comment and he fails horribly at humor.  \\[0.6em]

\hrule \vspace{0.6em}

\textbf{Polyjuice:} \\
He didn’t take responsibility for his comment and fails horribly at humor.  \\
He didn’t take responsibility for comment and he fails horribly at humor.  \\
He didn’t take responsibility and he fails horribly at humor.  \\
}

\textexamplenofig{
\textbf{You trade in a car and they sell it at a profit.} (FinQA) \\[0.6em]

\hrule \vspace{0.6em}

\textbf{Ours:} \\
They sell it.  \\
You trade in a car.  \\
They sell it at a profit.  \\
You trade in a car and they sell it.  \\
You trade in a car and they sell it at a profit.  \\[0.6em]

\hrule \vspace{0.6em}

\textbf{Polyjuice:} \\
You trade a car and they sell it at a profit.  \\
You trade in a car and sell it at a profit.  \\
}

\textexamplenofig{
\textbf{You’re losing money in more than one way on that investment.} (FinQA) \\[0.6em]

\hrule \vspace{0.6em}

\textbf{Ours:} \\
You’re losing money.  \\
You’re losing money in way.  \\
You’re losing money on that investment.  \\
You’re losing money in than one way.  \\
You’re losing money in more than one way.  \\
You’re losing money in way on that investment.  \\
You’re losing money in than one way on that investment.  \\
You’re losing money in more than one way on that investment.  \\[0.6em]

\hrule \vspace{0.6em}

\textbf{Polyjuice:} \\
You’re losing money in one way on that investment.  \\
You’re losing money in more than one way that investment.  \\
You’re losing money in more than one way.  \\
}

\textexamplenofig{
\textbf{The most promising agents in clinical development are reviewed.} (TinyTextbooks) \\[0.6em]

\hrule \vspace{0.6em}

\textbf{Ours:} \\
The agents are reviewed.  \\
The promising agents are reviewed.  \\
The most promising agents are reviewed.  \\
The agents in development are reviewed.  \\
The promising agents in development are reviewed.  \\
The agents in clinical development are reviewed.  \\
The most promising agents in development are reviewed.  \\
The promising agents in clinical development are reviewed.  \\
The most promising agents in clinical development are reviewed.  \\[0.6em]

\hrule \vspace{0.6em}

\textbf{Polyjuice:} \\
The most promising agents in clinical development.  \\
Most promising agents are reviewed.  \\
The most promising agents are reviewed.  \\
The most promising agents in development are reviewed.  \\
}

\textexamplenofig{
\textbf{For all kinds of business problems, we are there to help you to resolve business problems by astrology.} (TinyTextbooks) \\[0.6em]

\hrule \vspace{0.6em}

\textbf{Ours:} \\
We are.  \\
We are there.  \\
For all kinds, we are.  \\
For all kinds, we are there.  \\
We are to help you to resolve business problems.  \\
For all kinds of business problems, we are.  \\
We are there to help you to resolve business problems.  \\
We are to help you to resolve business problems by astrology.  \\
For all kinds of business problems, we are there.  \\
For all kinds, we are to help you to resolve business problems.  \\
We are there to help you to resolve business problems by astrology.  \\
For all kinds, we are there to help you to resolve business problems.  \\
For all kinds, we are to help you to resolve business problems by astrology.  \\
For all kinds of business problems, we are to help you to resolve business problems.  \\
For all kinds, we are there to help you to resolve business problems by astrology.  \\
For all kinds of business problems, we are there to help you to resolve business problems.  \\
For all kinds of business problems, we are to help you to resolve business problems by astrology.  \\
For all kinds of business problems, we are there to help you to resolve business problems by astrology.   \\[0.6em]

\hrule \vspace{0.6em}

\textbf{Polyjuice:} \\
For all kinds of business problems, we are there to help you to resolve problems by astrology.  \\
For all kinds of problems, we are there to help you to resolve business problems by astrology.  \\
For business problems, we are there to help you to resolve business problems by astrology.  \\
For all problems, we are there to help you to resolve business problems by astrology.  \\
For all kinds problems, we are there to help you to resolve business problems by astrology.  \\
For all kinds of business problems, we are there to help you resolve business problems by astrology.  \\
}

\section{Polyjuice Experiments}
\label{sec:polyjuice_why_not_good}

Polyjuice generates perturbations by sampling from a conditioned language model with beam search.
This is a stochastic process with a distribution for each iteration.
Because the distribution remains the same, continuous sampling from it yields mostly the same outputs.
Thus, one needs to sample increasingly more to obtain unique outputs.
These unique outputs are found on the tail of the distribution.
At the same time, such low-probability samples also tend to have lower fluency because of language modeling bias towards fluent language.

\begin{figure}[htbp]
    \centering
    \includegraphics[width=\linewidth]{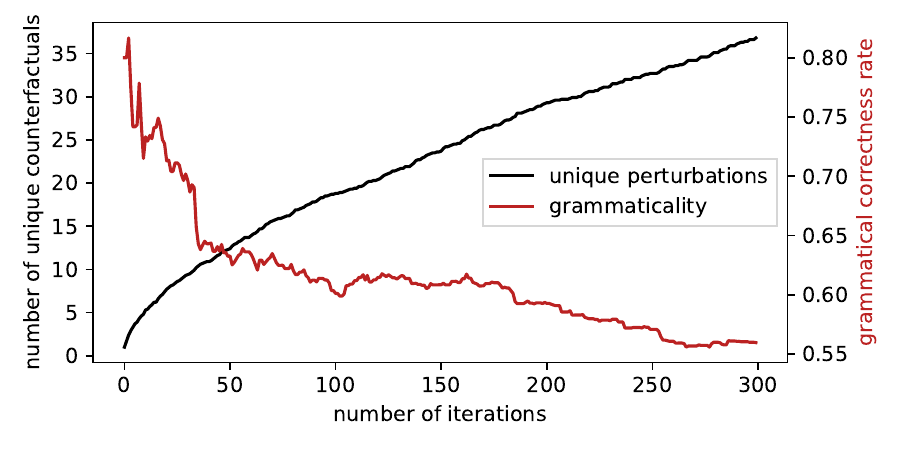}
    \caption{Number of observed unique perturbations and their estimated grammaticality (following \Cref{sec:experiment}) with respect to the increasing number of samplings from Polyjuice (up to 300). The values are averaged across the 10 texts from \Cref{sec:perturbation_examples}. Unique perturbations get more scarce and their grammaticality plummets.}
    \label{fig:polyjuice_inf_runtime}
\end{figure}

In our case, this translates to only a few unique observed perturbations with the grammaticality rate going down.
We demonstrate this by running 50 iterations for each of the 10 examples from \Cref{sec:perturbation_examples} and show the results in \Cref{fig:polyjuice_inf_runtime}.

\section{The Prediction Task in the User Study}
\label{sec:user-study-design}

\begin{figure}[htb]
    \includegraphics[width=\linewidth]{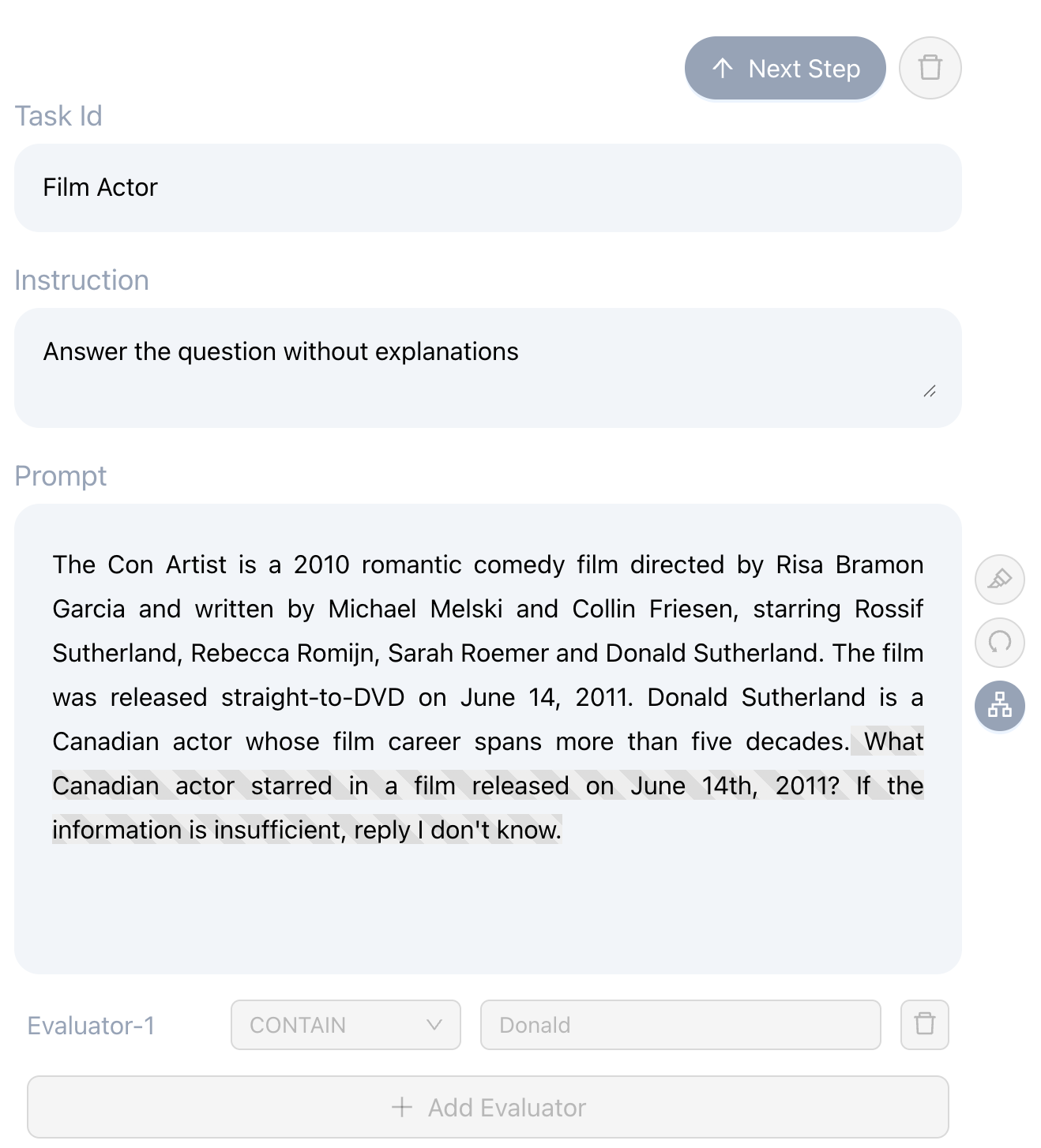}
    \caption{The prediction task used in the user study.}
    \label{fig:user-study-task}
\end{figure}

\begin{figure*}[htb]
    \includegraphics[width=\linewidth]{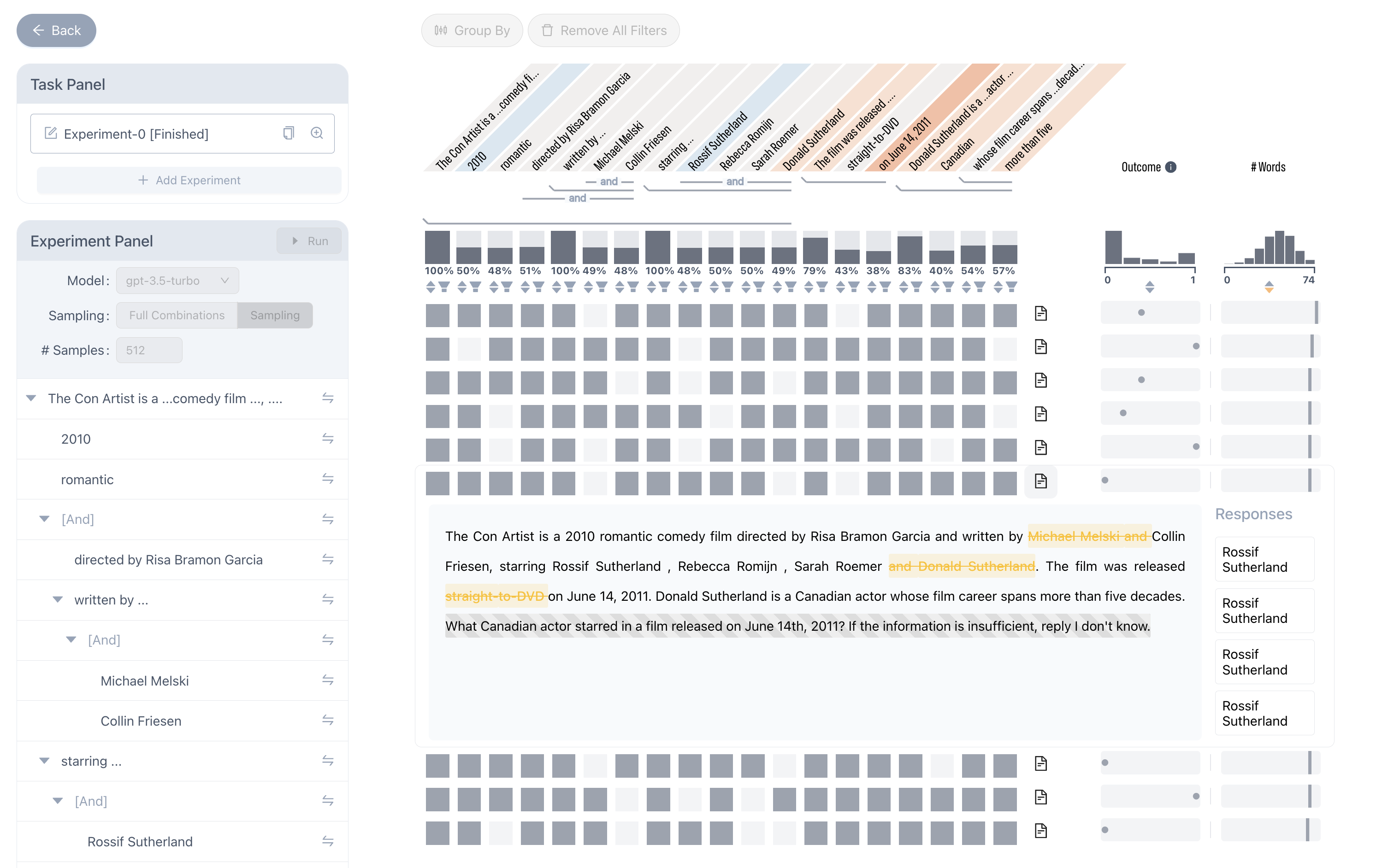}
    \caption{The interface viewed by users during the user study. }
    \label{fig:user-study-example}
\end{figure*}

During the user study, the participants are required to \re{analyze}{analyzer} GPT 3.5's behavior in solving the task shown in \Cref{fig:user-study-task}. 
This task is selected from the HotpotQA dataset~\cite{yang2018hotpotqa}, which is used to test the model's multi-step reasoning capability.  
In the task, the model has to find the answer of ``What Canadian actor starred in a film released on June 14th, 2011?'' 
The correct answer is ``Donald Sutherland''. 
So we define an evaluator defined as whether the model's response contains the word ``Donald'', which distinguish the correct answers with incorrect ones. 
To answer this question \re{correctly}{collectly}, the model needs to capture three piece of information, ``The Con Artist is a movie released in June 14, 2011''; ``Donald Sutherland acted in this movie''; and ``Donald Sutherland is Canadian.''
To avoid the model \re{using}{uses} its own knowledge to find the answer, we add a constraint that ``if the information is insufficient, reply I don't know.''
This newly added sentence and the question sentence are always included in all counterfactuals without any perturbations (marked in the dashed area). 
\Cref{fig:user-study-example} shows how the visualization of the counterfactuals from this example.

\re{\section{Estimiting the Shapley Values using KernelSHAP}
\label{sec:kernelshap}
In this section, we introduce the algorithm we used to calculate the feature attributions visualized in table header. 

Using the segmentation algorithm described in Section~4, we represent each perturbed text instance as binary vectors $z^{(1)}, \dots, z^{(N)} \in \{0,1\}^M$, where each entry denotes the presence (1) or absence (0) of a specific text segment. 
Each perturbation is passed through the LLM and the evaluator to obtain $y_j = f(z^{(j)}) \in [0, 1]$, denoting the the probability that the evaluator condition is satisfied. 
This allows us to frame the task as a local explanation problem: we aim to estimate additive attributions $\{ \phi_i \}$ such that the model can be approximated by a linear surrogate of the form
\[
f(z) \approx \phi_0 + \sum_{i=1}^M \phi_i z_i.
\]
To compute these SHAP values, we implemented the KernelSHAP algorithm proposed by Lundberg and Lee~\cite{lundberg2017shap}.
In KernelSHAP, the goal is to efficiently approximate Shapley values by solving a weighted linear regression problem.
\[
\min_{\phi} \sum_{z \in \{0,1\}^M} w(z) \left[ f(z) - \phi_0 - \sum_{i=1}^{M} \phi_i z_i \right]^2
\]
Each perturbation gets a weight $w(z^{(j)})$ that reflects how important it is in approximating the true Shapley value. 
The kernel weight is defined as:
\[
\omega(z^{(j)}) = \frac{M - 1}{\binom{M}{|z^{(j)}|} \cdot |z^{(j)}| \cdot (M - |z^{(j)}|)},
\]
where $|z^{(j)}|$ is the number of preserved segments. 

\noindent
This weighed linear regression problem has a close-form solution
\[
    \begin{pmatrix} \hat{\phi}_0 \\ \hat{\boldsymbol{\phi}} \end{pmatrix}
    = (\tilde{\mathbf{X}}^\top \mathbf{W} \tilde{\mathbf{X}})^\dagger \tilde{\mathbf{X}}^\top \mathbf{W} \mathbf{y},
\]
where $\mathbf{W} = \mathrm{diag}(\omega_1, \dots, \omega_N)$, $\mathbf{X} \in \{0,1\}^{N \times M}$ is the perturbation matrix and the $\mathbf{y} \in [0, 1]^N$ is the prediction vector, and $(\cdot)^\dagger$ denotes the Moore–Penrose pseudoinverse for numerical stability.
\smallskip

By solving this problem, we gain ${\hat{\phi}_i}$ as an approximation of the Shapley value that quantifies the contributions of each segment.}

\end{document}